\RequirePackage{rotating}

\documentclass[sigconf, nonacm]{acmart}

\usepackage{paralist}
\usepackage{multirow}
\usepackage{array}
\usepackage{booktabs}
\newcolumntype{M}[1]{>{\centering\arraybackslash}m{#1}}
\usepackage{geometry}
\usepackage{rotating}
\usepackage{makecell}
\usepackage{enumitem}
\usepackage{textcmds}
\usepackage{xspace}
\usepackage{balance}

\newcommand{\numberTools}{36\xspace}

\begin{document}
\title{Navigating Tabular Data Synthesis Research}
\subtitle{Understanding User Needs and Tool Capabilities}

\author{Maria F. Davila R.}
\affiliation{%
  \institution{OFFIS - Institute for Informatics}
  \city{Oldenburg}
  \country{Germany}
  \postcode{26121}
}
\email{maria.davila@offis.de}

\author{Sven Groen}
\affiliation{%
  \institution{initions GmbH}
  \city{Hamburg}
  \country{Germany}
}
\email{sven.groen@initions-consulting.com}

\author{Fabian Panse}
\affiliation{%
  \institution{Hasso Plattner Institute}
  \city{Potsdam}
  \country{Germany}
}
\email{fabian.panse@hpi.de}

\author{Wolfram Wingerath}
\affiliation{%
  \institution{Carl von Ossietzky University of Oldenburg}
  \city{Oldenburg}
  \country{Germany}
}
\email{wolfram.wingerath@uol.de}

\begin{abstract}
In an era of rapidly advancing data-driven applications, there is a growing demand for data in both research and practice. Synthetic data have emerged as an alternative when no real data is available (e.g., due to privacy regulations). Synthesizing tabular data presents unique and complex challenges, especially handling 
\begin{inparaenum}[(i)]
\item missing values, 
\item dataset imbalance, 
\item diverse column types, and 
\item complex data distributions,
\end{inparaenum}
as well as preserving
\begin{inparaenum}[(i)]
\item column correlations, 
\item temporal dependencies, and
\item integrity constraints (e.g., functional dependencies)
\end{inparaenum}
present in the original dataset. While substantial progress has been made recently in the context of generational models, there is no one-size-fits-all solution for tabular data today, and choosing the right tool for a given task is therefore no trivial task. In this paper, we survey the state of the art in Tabular Data Synthesis (TDS), examine the needs of users by defining a set of functional and non-functional requirements, and compile the challenges associated with meeting those needs. In addition, we evaluate the reported performance of \numberTools popular research TDS tools about these requirements and develop a decision guide to help users find suitable TDS tools for their applications. The resulting decision guide also identifies significant research gaps.
\end{abstract}

\maketitle

\setlength{\parskip}{2pt}
\section{Introduction}
\label{sec:intro}

The efficacy of data-driven models is limited by the accessibility of sufficient and diverse data. One of the main reasons for this lack of data is privacy regulations, which apply in particular to person-related data such as medical data or movement profiles. Other reasons are expensive or time-consuming data collection processes or the fact that scenarios are to be considered that have hardly ever existed before. Synthetic data emerge as a solution \cite{Le} for this problem. Data synthesis is generating realistic artificial data that accurately replicates the characteristics of real datasets. The generation of synthetic data gained new momentum with the success of deep learning techniques in image~\cite{Dhariwal}, text~\cite{Radford} and table generation~\cite{DDPM, TableGAN, Xu}.

Our study focuses on tabular data, which is organized into rows (or records) representing individual data points, and columns representing different features of those data points. According to our research, there is currently no tabular data synthesis (TDS) tool that works well across all applications and there is also no benchmark to measure a TDS tool's \qq{fitness for use}, which refers to the capability of the tool to meet the specific functional and non-functional requirements and demands of a certain application. Selecting a TDS tool for a specific use case is therefore a major challenge. Although commercial platforms, such as the Synthetic Data Vault (SDV)~\cite{SDV} or Gretel AI~\cite{Gretel}, choose and adapt TDS tools, they do not validate their suitability for specific use cases. Moreover, we focus on data-driven synthesis, which receives a real dataset as input and generates a synthetic dataset as output based on the input data. 
This means that approaches generating data primarily based on a given schema and some additional information, such as domains or statistics, 
(e.g., ~\cite{NeufeldML93,GraySEBW94,BrunoC05,HoukjaerTW06,HoagT07,ChristenP09,RablP11})
are not the focus of this paper.

To ensure the best possible utility of the generated data, a universal tool must fully capture and replicate several characteristics of the original data, including column types and distributions, correlations between columns, and integrity constraints, such as unique constraints or functional dependencies. In addition, the resource requirements of this use case must be met in terms of runtime and hardware requirements. This list of characteristics can be seen as functional and non-functional requirements that TDS tools must fulfill, in order to be suitable for specific use cases. 

The main goal of our research is to provide the basis for the evaluation of the suitability of TDS tools for use-case specific requirements. Our contributions are:
\begin{itemize}[leftmargin=*, itemsep=0pt, topsep=0pt]
    \item[$c_1$] A \textbf{survey} of current research on data-driven TDS, 
   \item[$c_2$] The definition of \textbf{functional} and \textbf{non-functional requirements} that allow to assess a TDS tool's suitability for a specific use case,  
   \item[$c_3$] An \textbf{assessment} of \numberTools tools with respect to their reported performance on those requirements, and
    \item[$c_4$] A \textbf{decision guide} for users searching for a suitable TDS tool for their use case, developed by compiling the reported performance of the leading TDS tools on such requirements. 
\end{itemize}

The remainder of the paper is structured as follows. Section~\ref{sec:related} presents related work. In Section~\ref{sec:purpose}, we discuss the main purposes for TDS where we base our discussion on often-quoted reasons why users need to create synthetic tabular data. Section~\ref{sec:tab_challenges} names the main challenges in TDS and its differences to image and text generation. These two sections are the basis for the identification of the functional and non-functional requirements, which we present in Section~\ref{sec:requirements}. While Section~\ref{sec:background} describes our assessment of the capabilities of predominant TDS tools, Section~\ref{sec:evaluation} gives a brief overview of the possible evaluation techniques for synthetic tabular data. This results in our assessment matrix and decision guide presented in Section~\ref{sec:guide}. Finally, Section~\ref{sec:conclusion} concludes the paper and gives an outlook on our upcoming research.

\section{Related Work}
\label{sec:related}

The Synthetic Data Vault (SDV) \cite{SDV}, Gretel AI \cite{Gretel}, and Mostly AI \cite{Mostly} are platforms for the generation of tabular data. These platforms must choose from the available tools to address the widest range of use cases possible. However, these platforms do not report on the specific limitations of those tools. In contrast, our goal is to create a framework that allows to identify use-case specific requirements and determine those limitations.

Several surveys served as input to our work to identify the predominant approaches and TDS tools. Hernandez et al. \cite{Hernandez}, Fan et al. \cite{Fan}, Figueira et al, \cite{Figueira}, and Brophy et al. \cite{Brophy} explored the use of Generative Adversarial Networks (GANs) for health records, categorical and numerical data types and time series generation. Koo and Kim \cite{Koo} reviewed generative diffusion models for tabular data, paralleling Lin et al. \cite{Lin}, who focused on time-series diffusion. Both identified three key methodologies: Denoising Diffusion Probabilistic Models (DDPM), Score-based Generative Models (SGM), and Stochastic Differential Equations (SDE). Fonseca and Bacao~\cite{FonsecaB23} recently provide an extensive survey on tabular data synthesis including an evaluation of 70 tools across six different machine learning problems. 

However, while our focus is on various approaches from the field of generative deep learning (including GANs, autoencoders, probabilisitc diffusion, graph neural networks, and transformers), the only deep learning approaches they consider are GANs and autoencoders. Moreover, they do not address the problem of finding the most suitable tool for a specific use case and therefore do neither define (non-)functional requirements for data synthesis nor evaluate their tools in terms of those requirements.

\begin{figure*}
    \centering
    \includegraphics[scale=0.82]{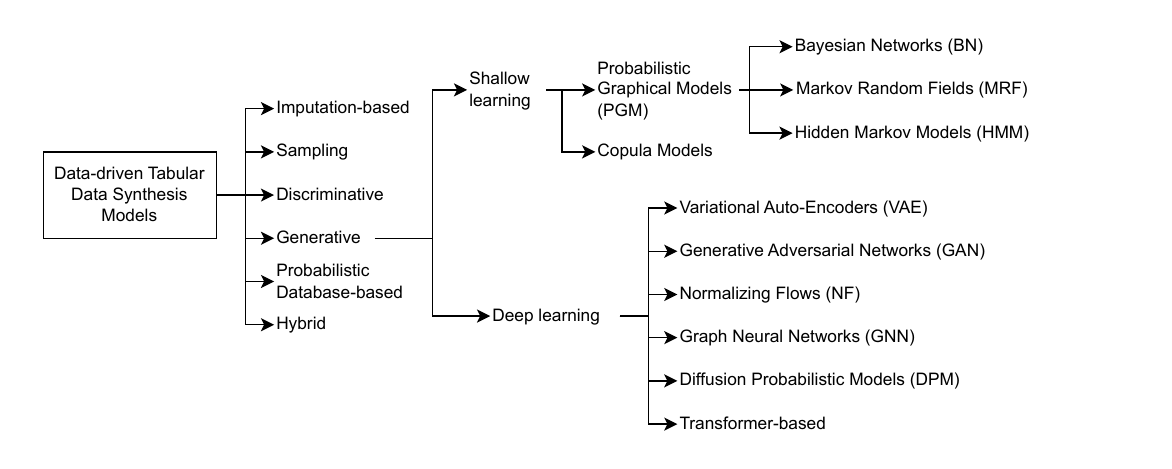}
    \Description[<TDS Models Classification>]{<Classification of the data-driven TDS models included in our study.>}
    \caption{Classification of the data-driven TDS models included in our study.}
    \label{img:methods}
\end{figure*}

To summarize, none of these surveys provide a comparison of deep learning approaches (see Figure~\ref{img:methods}) as we do in this paper. Additionally, they do not provide any insights into how users can assess a tool's fitness for use, or guide them in the process of choosing a suitable TDS tool for their specific use case. 

\section{Purposes of Tabular Data Synthesis}
\label{sec:purpose}

Based on our literature review, we pinpointed five purposes why users need to generate artificial tabular data:
\begin{itemize}[leftmargin=*, itemsep=0pt, topsep=0pt]
    \item \textbf{Missing Values Imputation:} Datasets often have incomplete entries, which can distort analyses. For instance, electronic health records (EHRs), where the patient's smoking status is missing, which is vital information for predicting the risk of heart disease. TDS allows users to generate datasets that are free from these gaps, ensuring intentional completeness~\cite{NaumannFL04}. 
    \item \textbf{Dataset Balancing:} Some datasets have a few classes which significantly outnumber others, risking bias towards these dominant classes. For example, in a diabetes dataset, non-diabetic patient records may outnumber diabetic ones. This discrepancy risks biasing predictive models towards the dominant non-diabetic class. TDS can rectify this imbalance by generating additional synthetic diabetic patient records. 
    \item \textbf{Dataset Augmentation:} TDS can be used for data augmentation, where the goal is to expand datasets for enhancing model robustness and generalization. In our EHR example, this would mean synthesizing records for new patients from all classes.
    \item \textbf{Customized Generation:} The generation of synthetic datasets must sometimes be directed by external factors, in order to create specific scenario data, particularly when original data does not exist. For instance, in the EHR context, researchers might need to generate data for a scenario where there's a spike in a particular disease due to an environmental change. 
    \item \textbf{Privacy Protection:} Many domains contain sensitive data, requiring effective measures to protect privacy when these data need to be shared or reused. In our EHR example, a hospital would like to share data for external analysis and therefore creates synthetic patient records that closely resemble real statistical patterns but do not correspond to actual patient data.
  
\end{itemize}

Privacy protection can be the sole reason for the synthesis but it can also be combined with any of the other purposes. While tools for class imbalance or missing values can most often be adapted for data augmentation or vice versa, it is important to choose the right one for the specific task at hand, as key relationships in the data might not be preserved. In addition it will likely increase the workload and computational costs.

\section{Tabular Data Synthesis Challenges}
\label{sec:tab_challenges}
For all domains of data synthesis, whenever privacy protection is of interest, one challenge is the \textit{Privacy vs. Utility trade-off} \cite{TableGAN}. Data utility refers to the ability of the data to serve its intended purpose effectively. Generating synthetic samples while preserving privacy is challenging because enhancing privacy often diminishes the utility of the data, and vice versa~\cite{Kamino}. The level of privacy can be achieved using Differential Privacy (DP), a rigorous mathematical framework for guaranteeing privacy in statistical analysis \cite{Dwork2006}.

TDS tools must capture and replicate the main characteristics of the original (real) dataset. This includes the column types and correlations between columns. This is challenging because of the following reasons:

\begin{itemize}[leftmargin=*, itemsep=0pt, topsep=0pt]
	\item \textbf{Missing values:} Accurately capturing the characteristics of the data with information gaps is challenging because these gaps represent information loss and the generated data could be a poor representation of the correlations between columns.
	\item \textbf{Imbalanced datasets:} Capturing the characteristics of the minority classes is especially complicated, since these classes are underrepresented. As a consequence of this under-representation, some algorithms may over fit the data, suffer from phenomena such as \qq{mode collapse}, where some classes are not generated at all \cite{GAN}, or generate unrealistic samples of the minority classes~\cite{SMOTE}. However, generating accurate samples of these minority classes can be very crucial for many applications, such as intrusion detection~\cite{intrusion}, that aim to identify outliers.
	\item \textbf{Diversity of column types:} Unlike images, tabular datasets usually contain a mix of different column types, such as numerical, categorical, temporal, text, or even mixed types consisting of values from different basic types. Different column types might require distinct pre-processing or handling techniques.
	\item \textbf{Complex column distributions:} The distribution of a column contains its spread, tendencies, and patterns in the data, providing valuable insights into its characteristics and relationships with other columns. Capturing complex distributions is challenging because traditional methods such as simply modeling mean and standard deviation may not be sufficient to characterize non-Gaussian distributions. 
	\item \textbf{Temporal Dependencies:} The temporal dimension of time series data introduces an additional layer of complexity. Two particular challenges for time series generation are discrete time series, because backpropagation presents problems \cite{Brophy}, and long-term dependencies, because their discovery and modeling require extra memory \cite{DoppelGANger}. 
\end{itemize}

\section{User Needs: TDS Requirements}
\label{sec:requirements}
Using the identified purposes in Section \ref{sec:purpose} and challenges in Section \ref{sec:tab_challenges}, we have determined a list of functional and non-functional requirements shown in Table~\ref{tab:requirements}. Current TDS tools differ from each other in what characteristics they are able to capture and replicate. We have identified 12 requirements.

\begin{table*}[t]
  \caption{Functional and Non-functional requirements for TDS Tools}
  \label{tab:requirements}
  \centering
  \scalebox{1}{
  \begin{tabular}{|p{2cm}|p{5cm}|p{8cm}|}
    \hline
      & \textbf{Requirement} & \textbf{Possible Categories} \\
    \hline
    \multirow{6}{*}[-5em]{\textbf{Functional}} & Ability to work with multiple non-independent columns & Single column, two columns, or multi-column datasets. \\
    \cline{2-3}
    & Ability to handle different types of columns effectively & Categorical, numerical (continuous and discrete), temporal, text, and mixed (e.g., categorical and numerical). \\
    \cline{2-3}
    & Ability to accurately capture and replicate the real univariate column distribution & Gaussian and other typical statistical distributions (uniform, exponential, Poisson, binomial, Logistic, etc.), skewed, multinomial. \\
    \cline{2-3}
    & Ability to preserve correlations between columns &
    Joint and conditional probabilities of subsets of columns.\\
    \cline{2-3}
    & Ability to preserve temporal dependencies between columns & Short-term and long-term dependencies. \\
    \cline{2-3}
    & Ability to preserve the integrity constraints of the dataset & Rules or conditions enforced. They can be intra-record (only one record) or inter-record, such as unique column combinations (UCCs), functional dependencies (FDs), inclusion dependencies (INDs), or denial constraints (DCs). \\
    \cline{2-3}
    & Ability to preserve the inter-table correlations & Parent-child relations, or relationships between multiple tables. \\
    \hline
    \multirow{5}{*}[-5em]{\textbf{Non-functional}} & Level of configuration the tool needs & Represents the ability of the tool to synthesize datasets without the need for extensive configuration or fine-tuning. It represents its "out-of-the-box" capability. \\
    \cline{2-3}
    & Level of pre-processing the tool needs & Represents the need for pre-processing the input data. For example, handling missing values, normalizing columns to similar scales and ranges to support convergence, or encoding columns into a format that can be effectively processed by the model.\\
    \cline{2-3}
    & Hardware the tool needs & Represents the technical requirements of the TDS, including the need for a GPU for training. \\
    \cline{2-3}
    & Resource efficiency of the tool (time and memory) & Represents the time and memory required to synthesize a dataset. \\
    \cline{2-3}
    & Scalability of the tool & Represents the ability to efficiently handle increasingly large datasets while maintaining high performance and accuracy. \\
    \hline
  \end{tabular}}
\end{table*}

The first four functional requirements address 
\begin{inparaenum}[(i)]
\item the number of non-independent columns the tool can synthesize, 
\item the types and 
\item distributions of columns the tool is able to handle as well as
\item  whether the tool preserves the correlations between columns. 
\end{inparaenum}
Early attempts of data synthesis started with basic statistical models, random sampling, and rule-based approaches \cite{Brandt}, which were only able to generate one or two dependent columns at the same time, or they would focus either on only categorical or numerical columns \cite{SMOTE}. Similarly, early attempts of data synthesis used simplified models that assumed all columns would follow a Gaussian distribution. All the TDS tools included in this study are able to work with multi-column datasets, yet they differ on what column types, distributions, and correlations they can capture and replicate. 

The fifth functional requirement addresses temporal dependencies. Leading TDS tools, which address complex column distributions and correlations \cite{CTABGAN, Kotelnikov, REaLTabFormer}, do not address temporal dependencies. TDS tools especially designed for time series datasets differ on whether or how well they can preserve short or long-term dependencies.

The requirement to preserve integrity constraints refers to the ability of a TDS tool to create synthetic records, that do not violate the rules enforced on the dataset, such as unique constraints or functional dependencies~\cite{Kamino}. Finally, inter-table correlations refer to relations between columns that are preserved, even though the columns belong to different tables. 

The group of non-functional requirements addresses factors that are not directly related to how well the TDS tools capture and replicate a dataset, instead, they refer to operational characteristics. To compare different TDS tools, it is important to include factors such as
\begin{inparaenum}[(i)]
\item how much configuration is needed before the tools can be properly used, 
\item how much pre-processing of the input data is required so that the tools can process them, 
\item what hardware (e.g., GPUs) is required for executing the tools,
\item how much resources (e.g., runtime, memory, electrical power) the tools need, and
\item how well the tools scale with larger datasets.
\end{inparaenum}

\section{TDS Tool Capabilities}
\label{sec:background}
We adopt the classification of TDS models into process-driven and data-driven models as introduced in \cite{Goncalves} and extend it with sub-categories of data-driven models as in Figure \ref{img:methods}. Our focus is on data-driven TDS tools, which synthesize data using an existing (real-world) dataset as input. 

\subsection{Imputation-based Models}
\label{sec:imputation}
Imputation-based data synthesis models were initially introduced to reduce the risk of disclosing sensitive data \cite{Rubin2}, as solutions for Statistical Disclosure Control (SDC) and Statistical Disclosure Limitation (SDL). The most popular imputation-based TDS tools use either multiple imputation or masking techniques.

Multiple imputation is originally a model to handle missing values, where each missing value is replaced by two or more imputed values \cite{Rubin1}. Raghunathan et al. \cite{Raghunathan} expanded it for data synthesis, which is easy to understand and implement. Moreover, it does not require high computational resources. However, it is highly sensible to bias as it can produce either extreme samples or contain several repeats of the observed records~\cite{Reiter2002}.  

\subsection{Sampling Models}
Data synthesis is often used to balance datasets. The straightforward solution for this problem 
is the augmentation with data records of the minority class, known as \textit{random} over-sampling. Similarly, \textit{random} under-sampling consists of the removal of data records from the majority class. While \textit{random} over-sampling introduces an increased risk of over-fitting, \textit{random} under-sampling frequently results in the loss of valuable information intrinsic to the original dataset \cite{Brandt}. Tools like Synthetic Minority Over-Sampling Technique (SMOTE)~\cite{SMOTE} can be used for class imbalance but also to generate complete synthetic datasets. However its vanilla version only works for continuous data, and the synthesized records are linearly dependent on the original minority class records, often leading to over-fitting \cite{Brandt}. Variations of SMOTE address these limitations, such as borderline-SMOTE~\cite{borderSMOTE}, SVM-SMOTE~\cite{SVM-SMOTE}, K-Means-SMOTE~\cite{KMEANS-SMOTE}, and ADASYN~\cite{ADASYN}.

\subsection{Discriminative Models}
A distinction is made between generative and discriminative models. Discriminative estimate the conditional probability of the output given the input, in other words the probability of a label given an observation. However, they do not learn the interdependence between all the columns (target and non-target). 
Discriminative models can be leveraged for TDS using the learned conditional probabilities. In privacy-preserving data mining, clustering-based algorithms generate synthetic data while aiming to maintain certain properties of the original data, such as the distance-preserving maps from Liu et al.~\cite{Liu2008}. However, these models have limitations in handling more intricate data characteristics, because they are built to attempt to estimate the probability that an observation belongs to a class, and not to learn the complete distribution~\cite{Foster}.

\subsection{Generative Models}
Generative models aim to learn the joint probability distribution of the input and target columns, therefore they can generate the complete table with all its columns \cite{Goodfellow}. They are suitable for all the purposes introduced in Section \ref{sec:purpose} and can be classified into shallow and deep. 

\subsubsection{Shallow Generative Models}
Shallow generative models have simpler architectures with few or no layers of abstraction or transformation. We consider \textbf{Copula models} and \textbf{Probabilistic Graphical models} (PGMs) the most relevant for TDS. 

A \textbf{copula} is a mathematical function that links multivariate joint distribution functions to their one-dimensional marginal distribution functions~\cite{nelsen2006introduction}. Simply put, a copula separates the analysis of a multivariate distribution into two parts: the individual behavior of each variable, known as the marginals, and a function that binds the marginals back together, capturing how they relate with each other. The Gaussian Copula uses a multivariate normal distribution to model these relationships \cite{Li}. Tabular Copula~\cite{TabCopula} is a Python package that uses Gaussian Copulas to produce synthetic datasets that preserve the statistical properties of the original data. However, copulas, especially Gaussian copulas, can have problems capturing extreme dependency structures in data~\cite{Fermanian}.

On the other hand, \textbf{PGMs}~\cite{PGM} represent complex distributions through graphs where nodes represent variables and edges represent probabilistic dependencies between these variables. They enable inference as well as learning tasks and explicitly show the dependencies in the data. However, exact inference becomes computationally infeasible with large datasets. Popular PGMs in TDS can be subdivided into Bayesian networks, Hidden Markov Models, and Markov Random Fields.

\paragraph{\textbf{Bayesian Networks}}
(BN) are PGMs that represent a set of variables and their conditional dependencies via a directed acyclic graph (DAG). PrivBayes~\cite{PrivBayes} is a privacy-preserving TDS tool, which uses BNs and adds noise to the parameters of the model to ensure DP. However, it is reported to introduce unnecessary errors~\cite{PB-PGM}. BNs are also used in models of other classes (e.g., \cite{GANBLR}) to integrate semantic information on the input data into the learning process.

\paragraph{\textbf{Markov Random Fields}}
(MRF) are undirected graphical models, which can model complex interactions and dependencies without assuming a specific direction of influence. PrivMRF~\cite{PrivMRF} is a DP MRF tool that reports experiments producing synthetic data with better utility compared to other DP tools. PrivLava~\cite{PrivLava} is a tool that uses graphical models to capture complex correlations and aims to synthesize tabular data with complex schemas where multiple tables are connected via foreign keys under DP. It also reports better utility results for single tables than PrivBayes and PrivMRF but at the expense of higher computational costs. Experiments with respect to other functional requirements and metrics are not reported.

\paragraph{\textbf{Hidden Markov Models}}
(HMMs) are a type of PGM that uses a graphical representation to model the probabilistic relationships between a series of observed and hidden states, using a Markov process. This allows HMMs to compute joint probabilities across sequences. 

\subsubsection{Deep Generative Models}
\label{sec:deep_gens}

Deep Generative Models are composed of multiple layers that enable the model to learn hierarchical representations. They leverage deep learning techniques to model the joint probability distribution of a dataset. 

Based on our review, the most popular models are discussed in the sections below. 

\paragraph{\textbf{Variational Autoencoder}}
(VAEs)~\cite{Kingma} combine variational inference and autoencoders. The encoder network maps the input data to a latent space and the decoder network reconstructs the original input from this latent space. By sampling from the latent space, VAEs can synthesize new data records. 

VAEs have demonstrated impressive results in synthesizing data across multiple domains, including images, text, and music \cite{Doersch, Bowman, Roberts}. In TDS, VAEs preserve the characteristics of the dataset better than sampling or shallow approaches \cite{Xu}. Nevertheless, VAEs have limitations. One notable challenge is their propensity to over-simplify the complex, real-world distributions inherent in the original data due to the standard Gaussian assumption imposed on the latent space \cite{Nalisnick}. This might lead to synthetic data that simplifies the original data. Additionally, VAEs can struggle with discrete or categorical columns because they use a reparametrization trick for backpropagation, which only works well for continuous latent spaces \cite{Jang}.

Some important tools include discrete VAE~\cite{Rolfe}, DP-VAEGM~\cite{DPVAEGM}, and tabular VAE (TVAE)~\cite{Xu}, which address these limitations, extending the range and improving the quality of the synthesized data. TimeVAE~\cite{TimeVAE} is able to generate multivariate time series data and its reported performance is similar to that of TimeGAN~\cite{TimeGAN}. 

\paragraph{\textbf{Generative Adversarial Networks}}
(GANs)~\cite{GAN} consist of two main neural networks: a generator and a discriminator. The generator uses random noise as input and generates synthetic data samples, while the discriminator aims to distinguish between real and synthetic data. During training, the generator and discriminator are trained in an adversarial manner, with the generator attempting to generate data that fools the discriminator, and the discriminator striving to correctly classify real and synthetic data. Through this competitive process, and using their implicit modeling of the data distribution, GANs learn to generate realistic and high-quality synthetic data samples that closely resemble the distribution of the real training data~\cite{CTABGAN}.

Tools such as medGAN~\cite{MedGAN}, DP-GAN~\cite{DPGAN}, and PATE-GAN~\cite{PATE-GAN} were specifically developed for privacy-preserving data synthesis. However, they sacrifice data utility and report lower performance than a vanilla GAN for machine learning tasks \cite{PATE-GAN}. TableGAN~\cite{TableGAN}, TGAN~\cite{XV18}, and CTGAN~\cite{Xu} are able to achieve high data privacy with better data utility. Building upon them, the two predominant GAN tools nowadays are CTAB-GAN~\cite{Zhao} and its successor CTAB-GAN+~\cite{CTABGAN}. They both work with mixed data types, imbalanced datasets, and complex distributions. 

GANBLR and its successor GANBLR++~\cite{GANBLR, GANBLR++} address the fact that GANs are not interpretable. C3-TGAN~\cite{C3TGAN} addresses that GANs do not consider explicit attribute correlations and property constraints.
Both approaches solve their respective problem by using Bayesian networks.

Most of the approaches for time series data use Recurrent Neural Networks (RNNs) \cite{RNN}, especially of the type Long Short-Term Memory (LSTM) \cite{LSTM}. TimeGAN \cite{TimeGAN} combines a GAN model with Autoregressive Models (AR) but it chunks the dataset into 24 epochs, which is not adequate for long-term dependencies~\cite{DoppelGANger}. DoppelGANger~\cite{DoppelGANger} is a custom workflow developed to address the key challenges of time series GAN approaches, such as long-term dependencies, complex multidimensional relationships, mode collapse, and privacy. DoppelGANger is currently used in commercial applications for synthetic data~\cite{Gretel}. 

\paragraph{\textbf{Normalizing Flows}}
(NF) use invertible and differentiable transformations to convert simple distributions like Gaussians into complex ones for probabilistic density modeling. This process is flexible and allows exact likelihood estimation but is computationally intensive. For this reason, there are not many NF TDS tools. 
Durkan et al.~\cite{Durkan} demonstrated the effectiveness of NF on tabular and image data. Yet, Manousakas et al.~\cite{usefulness} suggest that NF might underperform compared to models such as CTGAN~\cite{Xu} and TVAE~\cite{Xu}. Kamthe et al.~\cite{copulas} applied NF to learn the copula density for TDS, effectively capturing relations among columns. In most scenarios, however, it performs worse than TVAE \cite{Xu}.

\paragraph{\textbf{Graph Neural Networks}}
(GNNs) are specialized neural network architectures for processing graph-structured data. They excel in handling irregular data structures, using relationships between entities (nodes) and their connections (edges) in a graph. For TDS, rows are converted into graph nodes, connected by edges based on similarity or domain-specific knowledge. This transformation allows GNNs to learn node representations that capture inter-row and inter-column relations.

GOGGLE~\cite{Goggle} is a TDS tool integrating GNNs. By replacing typical VAE decoder architectures with GNNs, GOGGLE achieves realistic samples, highlighting the potential of GNNs in the synthesis of complex, domain-aware tabular data. However, GNNs can be computationally and memory-intensive, especially with large graphs~\cite{Iwata}. 

\paragraph{\textbf{Diffusion Probabilistic Models}}
(DPM) \cite{DDPM} are inspired by non-equilibrium physics and have gained significant importance with the improvements introduced by Yang et al.~\cite{Yang_DDPM} and Ho et al. \cite{DDPM2}. It involves a two-step training process, one forward diffusion step and a backward denoising step. DPMs model the data generation process as a reverse diffusion process, where noise is iteratively removed from a random initialization until a sample from the target distribution emerges~\cite{DDPM, DDPM2}. 

DPMs can be classified into three categories: Denoising Diffusion Probabilistic Models (DDPM), Score-based Generative Models (SGM), and Stochastic Differential Equations (SDE). They differ in how they transform noise into data records. DDPMs take a step-by-step approach, gradually refining noise. SGMs use the gradient of the data distribution to directly guide noise toward the outcome. Meanwhile, SDEs treat this transformation as a continuous process, modeling the addition and removal of noise through differential equations.

TabDDPM \cite{Kotelnikov} was the first highly cited TDS diffusion tool, and it is reported to outperform in terms of ML efficiency tools such as TVAE~\cite{Xu}, CTAB-GAN~\cite{Zhao}, CTAB-GAN+~\cite{CTABGAN}, and SMOTE ~\cite{SMOTE}. SOS~\cite{SOS} and STaSy~\cite{STaSy} are both SGMs that are reported to outperform TVAE~\cite{Xu} and TableGAN~\cite{TableGAN}. However, diffusion tools are computationally more expensive than other deep generative alternatives \cite{Lin}. 

TDS diffusion tools for time series data were initially used for time series forecasting,~\cite{TimeGrad, ScoreGrad, D3VAE} or to address missing values~\cite{CSDI, SSSD}. TSGM~\cite{TSGM} is a tool for multivariate time series generation, which uses an SGM model and generates records conditioned on past generated observations. 

\paragraph{\textbf{Transformer-based.}}
The Transformer architecture revolutionized the field of natural language processing (NLP) \cite{Vaswani}. It consists of an encoder-decoder structure and it replaces traditional RNNs and CNNs with self-attention mechanisms. This allows each element in a sequence to focus on different other elements of the same sequence, capturing long-range dependencies effectively. All current transformer-based tools use pre-trained large language models (LLMs), however, since other transformer architectures are also possible, we decided to call this category transformer-based.

Currently, only a few TDS approaches use transformers, with GReaT~\cite{GReaT}, REaLTabFormer~\cite{REaLTabFormer}, and TabuLa~\cite{TabuLa} being notable representatives. GReaT uses a pre-trained LLM (GPT-2) and transforms datasets into textual representations before providing them to the LLM (fine-tuning and inference). This step minimizes the pre-processing requirement. REaLTabFormer also uses GPT-2 and addresses the generation of synthetic datasets with two tables being in a one-to-many relationship (i.e., one parent and one child table). They aim to reduce extensive fine-tuning, especially for child tables. The authors of TabuLa emphasize that LLM-based TDS tools provide two main advantages: elimination of the need to pre-define column types and elimination of the dimension explosion problem when synthesizing high-dimensional data. However, LLM-based tools have limitations on training efficiency and preserving column correlations~\cite{TabuLa}.

\subsection{Probabilistic Database-based Models}
Ge et al.~\cite{Kamino} remark that the most popular (deep) generation models fail to preserve integrity constraints of the input data in the synthetic output data.
To address this issue, they developed a constraint-aware differentially private data synthesis approach called KAMINO.
KAMINO preserves denial constraints specified by the user.
Similar to VAEs KAMINO first uses the input dataset to learn a latent space and then uses this space to sample the synthetic dataset.
The difference is that KAMINO represents this space by a factorized probabilistic database~\cite{SaIKRR19}
and takes constraint violations into account when sampling the synthetic values one after another.
By learning weights for the individual constraints, 
KAMINO also allows the modeling of soft constraints, 
which do not strictly have to be fulfilled, but only to a certain extent.
Experiments show that KAMINO produces much fewer constraint violations
than other privacy-focused approaches, such as PrivBayes~\cite{PrivBayes}, DP-VAEGM~\cite{DPVAEGM}, and PATE-GAN~\cite{PATE-GAN}.
However, since KAMINO explicitly checks for constraint violations during sampling, its execution time is longer than that of these baseline approaches.

\subsection{Hybrid Models \& Other Approaches}

To improve their overall performance, some state-of-the-art TDS tools use combinations of different TDS models. Examples of such hybrid tools are AutoDiff~\cite{AutoDiff} and TabSyn~\cite{TabSyn}, which combine VAEs with a diffusion model to improve the performance of typical DDPMs for different column types and distributions. They are also designed to reduce runtime compared to typical diffusion tools. 

Gilad et al.~\cite{GiladPM21} proposed a method that uses already-generated tables and connects them via foreign keys while considering cardinality and integrity constraints. 
In combination with an existing TDS approach, this allows the generation of datasets with complex schemas.

\section{Synthetic Data Evaluation Metrics}
\label{sec:evaluation}
\begin{figure*}
    \centering
    \includegraphics[scale=0.82]{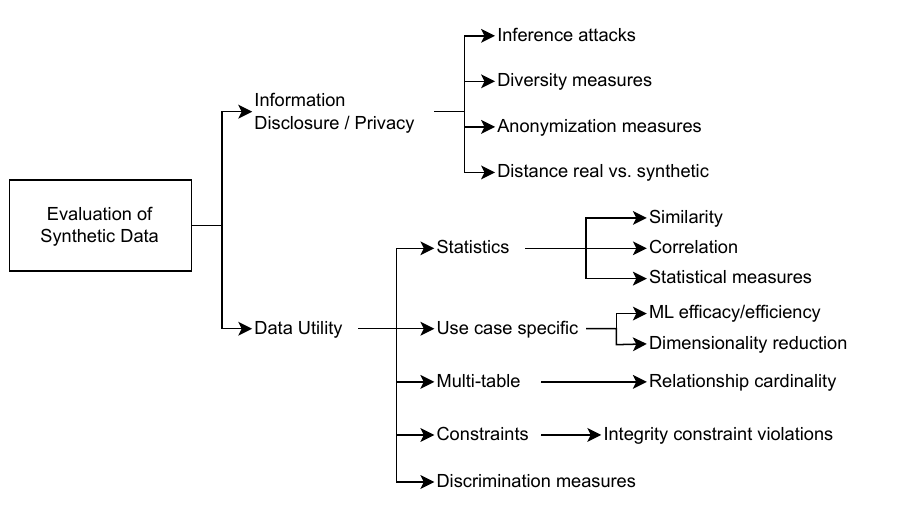}
    \Description[<Evaluation Classification>]{<Taxononomy of metrics for evaluating synthetic tabular data identified in our research>}
    \caption{Taxononomy of metrics for evaluating synthetic tabular data identified in our research.}
    \label{img:evaluation}
\end{figure*}

Evaluating the quality of synthetic tabular data is no trivial task for several reasons. The quality of other modalities of synthetic data, like images, can easily be identified by human inspection. This is not the case for tabular data because tabular datasets have inherent properties (see Section \ref{sec:tab_challenges}), such as their column distributions and correlations which are not easily recognizable by human inspection. 

There are numerous metrics proposed in the literature to measure the quality of synthetic tabular data, most of them capturing one certain characteristic of the data, e.g., whether its correlations are identical to the correlations in the input dataset. However, there is no universally accepted evaluation metric for synthetic data among researchers. This makes comparing the generative capabilities of the different approaches difficult, as each work uses its own set of metrics \cite{TABSYNDEX}.

First attempts, such as TabSynDex~\cite{TABSYNDEX}, aim to provide a universal metric by combining commonly used metrics into a single metric score.
However, we argue that a combined metric is only useful if its individual component metrics are appropriate for the purpose for which the synthetic dataset was created (Section~\ref{sec:purpose}).

Goncalves et al. \cite{Goncalves} classify TDS evaluation metrics into \qq{data utility} and \qq{information disclosure} metrics, coherent to the \textit{Privacy vs. Data Utility} trade-off discussed in Section \ref{sec:tab_challenges}.
Based upon this, we classify the different existing evaluation approaches into further, more detailed, classes.
Data utility metrics capture the usefulness of the synthetic dataset and how similar it is compared to its real counterpart. 
In contrast, information disclosure encompasses all metrics that are related to the privacy aspect of data synthesis.
Goncalves et al. describe them as measures of "(...) how much of the real data may be revealed (directly or indirectly) by the synthetic data" \cite[p. 6]{Goncalves}. In Figure~\ref{img:evaluation}, we present a comprehensive overview of evaluation metrics for synthetic tabular data in the form of a taxonomy. 

In Table~\ref{tab: Evaluation Metrics}, we present a selection of metrics (along with papers where they have been introduced or used) and classify them according to our taxonomy from Figure~\ref{img:evaluation}. In general, a metric should always be selected to suit the purpose of the considered synthesis task at hand. For example, when balancing the classes of a table, the goal is to change the distribution of some original columns. Thus, using a similarity-based metric for those columns is contradictory. 

\begin{table}[H]
\centering
\caption{Classification of evaluation metrics. The references are examples where the evaluation is employed, and they may not necessarily represent the first appearance of the evaluation techniques}
\label{tab: Evaluation Metrics}
\scalebox{0.95}{
\begin{tabular}{|l|l|}
\hline
\textbf{Class} &
\textbf{Evaluation Metric}   \\
\hline
Inference Attack & 
Categorical Correct Attribute\\ 
& Probability (CAP)~\cite{sdmetrics}, \\
& Membership Attack~\cite{Hernandez, TableGAN}  \\\hline
Anonymization & K-Anonymization~\cite{Synthcity},  \\
Measures & L-Diversity~\cite{Synthcity},  \\
& K-Map~\cite{Synthcity}  \\\hline
Distance & Novel Row Synthesis~\cite{sdmetrics}, \\
 Real vs. Synthetic & Common (leaked) Rows Proportion~\cite{Synthcity}, \\
& Distance to closest Record (DCR)~\cite{Hernandez,TableGAN}  \\
\hline
Similarity & Range Coverage~\cite{sdmetrics},\\
 & Outlier Coverage~\cite{sdmetrics},  \\
 & TV Complement~\cite{sdmetrics}  \\
& Kolmogorow-Smirnow-Test~\cite{KSTest, Bourou}, \\
 & Chi-squared Test~\cite{sdmetrics, Bourou}, \\
  & KL Divergence~\cite{Goncalves}, \\
  & Jensen-Shannon Distance~\cite{CTABGAN} \\
 & Contingency Similarity~\cite{sdmetrics} \\\hline
Statistical &  Mean, Median, Mode, \\
Measures  & Variance, Min, Max, \%-quantile \\\hline
Correlation & Pearson Coefficient~\cite{sdmetrics, Lu, Kotelnikov}, \\
& Spearman's Coefficient~\cite{sdmetrics,Bor} \\\hline
Discrimination & ML Real vs. Synthetic Discrimination~\cite{Synthcity, SENSEGEN}, \\
Measures & pMSE-score~\cite{PMSE, TABSYNDEX}  \\\hline
ML Efficacy & ML Classification~\cite{sdmetrics, Synthcity, Goncalves, Xu,Kotelnikov}, \\
& ML Regression~\cite{sdmetrics, Synthcity, Xu, Kotelnikov} \\\hline
Dimensionality & Principal Component Analysis (PCA)~\cite{TTSGAN,AIRGAN}, \\
Reduction & T-distributed stochastic neighbor \\
& embedding (T-SNE)~\cite{TTSGAN,AIRGAN} \\\hline
Relationship  & Cardinality Shape Similarity~\cite{sdmetrics}  \\
Cardinality  & \\\hline
IC Violations & g1-Error~\cite{Kamino} \\\hline
\end{tabular}
}
\end{table}

Additionally, it might be beneficial for beginners to consider using libraries such as the sdmetrics~\cite{sdmetrics} from the Synthetic Data Vault~\cite{SDV} or Synthcity~\cite{Synthcity}, as they offer comprehensive functionalities. These tools can be particularly helpful for those new to the field. The table serves as a starting point on where to look for a potential implementation.

In general, while creating synthetic tabular data, users need to carefully think about which aspects to focus on during evaluation and select their metrics accordingly.
This is especially crucial when handling sensitive data, where achieving strong metrics in privacy evaluation takes precedence over the overall data utility. This underscores the critical need for a careful and nuanced approach to evaluation, where the user strategically weighs and prioritizes dimensions based on contextual importance.

\section{Ability Assessment \& Decision Guide}
\label{sec:guide}
In our study, we assessed the \numberTools TDS tools listed in Table~\ref{tab:tools} on their suitability for the purposes described in Section \ref{sec:purpose} and their reported performance on the functional requirements listed in Table \ref{tab:requirements}. 
We have selected these tools based on their popularity, diversity, novelty, and quality (in terms of utility measures such as ML efficiency). The assessment resulted in the matrices shown in Tables~\ref{tab:matrix_pur} and \ref{tab:matrix_req}.

\begin{table}[ht]
    \centering
    \caption{The \numberTools TDS tools used in our assessment}
    \label{tab:tools}
    \scalebox{0.9}{
    \begin{tabular}{|l|l|}\hline
    \textbf{Model} & \textbf{TDS Tool} \\\hline
    Sampling & SMOTE~\cite{SMOTE}, Borderline-SMOTE~\cite{borderSMOTE}, \\
    & SVM-SMOTE~\cite{SVM-SMOTE}, Kmeans-SMOTE  \cite{KMEANS-SMOTE}, \\
    & ADASYN~\cite{ADASYN} \\
    \hline
    Bayesian Networks & PrivBayes \cite{PrivBayes} \\
    \hline
    MRF & PrivLava~\cite{PrivLava}, PrivMRF~\cite{PrivMRF}\\
    \hline
    GAN & medGAN~\cite{MedGAN}, PATE-GAN~\cite{PATE-GAN}, \\
    & DTGAN~\cite{DTGAN}, DP-GAN~\cite{DPGAN}, TableGAN~\cite{TableGAN},\\
    &  TGAN~\cite{XV18}, CTGAN~\cite{Xu}, C3TGAN~\cite{C3TGAN}\\
    & CTAB-GAN~\cite{CTABGAN}, CTAB-GAN+~\cite{Zhao},\\
    & GANBLR~\cite{GANBLR}, GANBLR++~\cite{GANBLR++}, \\
    & TimeGAN~\cite{TimeGAN}, DoppelGANger~\cite{DoppelGANger}\\
    \hline
    VAE & TVAE~\cite{Xu}, TimeVAE~\cite{TimeVAE}, DP-VAEGM~\cite{DPVAEGM}\\
    \hline
    Diffusion (DPM) & TabDDPM~\cite{Kotelnikov}, TSGM~\cite{TSGM},\\
    & SOS~\cite{SOS}, STaSy~\cite{STaSy} \\
    \hline
    Graph NN & GOGGLE~\cite{Goggle}\\
    \hline
    Transformer & GReaT~\cite{GReaT}, REalTabFormer~\cite{REaLTabFormer},\\
    &  TabuLa~\cite{TabuLa} \\
    \hline
    Prob. Database & KAMINO~\cite{Kamino}\\
    \hline
    Hybrid & AutoDiff~\cite{AutoDiff}, TabSyn~\cite{TabSyn}\\
    \hline
    \multicolumn{2}{c}{}\\[-0.5em]
    \end{tabular}
    }
\end{table}

\begin{table*}[p]
\centering
\caption{Assessment of the \numberTools TDS tools included in this study on their suitability for the different purposes.}
\label{tab:matrix_pur}
\scalebox{0.66}{
\begin{tabular}{l|cccccc|}
\cline{2-7}
 & \multicolumn{6}{c|}{\textbf{Purpose}} \\ \cline{2-7} 
 & \multicolumn{1}{c|}{\multirow{2}{*}{\textbf{Value Imputation}}} & \multicolumn{1}{c|}{\multirow{2}{*}{\textbf{Dataset Balancing}}} & \multicolumn{1}{c|}{\multirow{2}{*}{\textbf{Dataset Augmentation}}} & \multicolumn{2}{c|}{\textbf{Privacy}} & \multirow{2}{*}{\textbf{Customized}} \\ \cline{1-1} \cline{5-6}
\multicolumn{1}{|l|}{\textbf{Tools}} & \multicolumn{1}{c|}{} & \multicolumn{1}{c|}{} & \multicolumn{1}{c|}{} & \multicolumn{1}{c|}{\textbf{Simple}} & \multicolumn{1}{c|}{\textbf{Differential}} &  \\ \hline
\multicolumn{1}{|l|}{All SMOTE Variations} & \multicolumn{1}{c|}{} & \multicolumn{1}{c|}{$\times$} & \multicolumn{1}{c|}{} & \multicolumn{1}{c|}{} & \multicolumn{1}{c|}{} &  \\ \hline
\multicolumn{1}{|l|}{ADASYN} & \multicolumn{1}{c|}{} & \multicolumn{1}{c|}{$\times$} & \multicolumn{1}{c|}{} & \multicolumn{1}{c|}{} & \multicolumn{1}{c|}{} &  \\ \hline
\multicolumn{1}{|l|}{PrivBayes} & \multicolumn{1}{c|}{} & \multicolumn{1}{c|}{} & \multicolumn{1}{c|}{} & \multicolumn{1}{c|}{} & \multicolumn{1}{c|}{$\times$} &  \\ \hline
\multicolumn{1}{|l|}{PrivMRF} & \multicolumn{1}{c|}{$\times$} & \multicolumn{1}{c|}{$\times$} & \multicolumn{1}{c|}{$\times$} & \multicolumn{1}{c|}{} & \multicolumn{1}{c|}{$\times$} &  \\ \hline
\multicolumn{1}{|l|}{PrivLava} & \multicolumn{1}{c|}{$\times$} & \multicolumn{1}{c|}{$\times$} & \multicolumn{1}{c|}{$\times$} & \multicolumn{1}{c|}{} & \multicolumn{1}{c|}{$\times$} & $\times$ \\ \hline
\multicolumn{1}{|l|}{medGAN} & \multicolumn{1}{c|}{} & \multicolumn{1}{c|}{} & \multicolumn{1}{c|}{} & \multicolumn{1}{c|}{$\times$} & \multicolumn{1}{c|}{} &  \\ \hline
\multicolumn{1}{|l|}{PATE-GAN} & \multicolumn{1}{c|}{} & \multicolumn{1}{c|}{} & \multicolumn{1}{c|}{} & \multicolumn{1}{c|}{} & \multicolumn{1}{c|}{$\times$} &  \\ \hline
\multicolumn{1}{|l|}{DP-GAN} & \multicolumn{1}{c|}{} & \multicolumn{1}{c|}{} & \multicolumn{1}{c|}{} & \multicolumn{1}{c|}{} & \multicolumn{1}{c|}{$\times$} &  \\ \hline
\multicolumn{1}{|l|}{TableGAN} & \multicolumn{1}{c|}{$\times$} & \multicolumn{1}{c|}{$\times$} & \multicolumn{1}{c|}{$\times$} & \multicolumn{1}{c|}{$\times$} & \multicolumn{1}{c|}{} &  \\ \hline
\multicolumn{1}{|l|}{TGAN} & \multicolumn{1}{c|}{$\times$} & \multicolumn{1}{c|}{$\times$} & \multicolumn{1}{c|}{$\times$} & \multicolumn{1}{c|}{$\times$} & \multicolumn{1}{c|}{} &  \\ \hline
\multicolumn{1}{|l|}{DTGAN} & \multicolumn{1}{c|}{$\times$} & \multicolumn{1}{c|}{$\times$} & \multicolumn{1}{c|}{$\times$} & \multicolumn{1}{c|}{} & \multicolumn{1}{c|}{$\times$} &  \\ \hline
\multicolumn{1}{|l|}{CTGAN} & \multicolumn{1}{c|}{$\times$} & \multicolumn{1}{c|}{$\times$} & \multicolumn{1}{c|}{$\times$} & \multicolumn{1}{c|}{$\times$} & \multicolumn{1}{c|}{} & $\times$ \\ \hline
\multicolumn{1}{|l|}{C3TGAN} & \multicolumn{1}{c|}{$\times$} & \multicolumn{1}{c|}{$\times$} & \multicolumn{1}{c|}{$\times$} & \multicolumn{1}{c|}{} & \multicolumn{1}{c|}{} & $\times$  \\ \hline
\multicolumn{1}{|l|}{CTAB-GAN} & \multicolumn{1}{c|}{$\times$} & \multicolumn{1}{c|}{$\times$} & \multicolumn{1}{c|}{$\times$} & \multicolumn{1}{c|}{$\times$} & \multicolumn{1}{c|}{} & $\times$ \\ \hline
\multicolumn{1}{|l|}{CTAB-GAN+} & \multicolumn{1}{c|}{$\times$} & \multicolumn{1}{c|}{$\times$} & \multicolumn{1}{c|}{$\times$} & \multicolumn{1}{c|}{} & \multicolumn{1}{c|}{$\times$} & $\times$ \\ \hline
\multicolumn{1}{|l|}{GANBLR} & \multicolumn{1}{c|}{$\times$} & \multicolumn{1}{c|}{$\times$} & \multicolumn{1}{c|}{$\times$} & \multicolumn{1}{c|}{} & \multicolumn{1}{c|}{} & $\times$ \\ \hline
\multicolumn{1}{|l|}{GANBLR++} & \multicolumn{1}{c|}{$\times$} & \multicolumn{1}{c|}{$\times$} & \multicolumn{1}{c|}{$\times$} & \multicolumn{1}{c|}{} & \multicolumn{1}{c|}{} & $\times$ \\ \hline
\multicolumn{1}{|l|}{TimeGAN} & \multicolumn{1}{c|}{$\times$} & \multicolumn{1}{c|}{$\times$} & \multicolumn{1}{c|}{$\times$} & \multicolumn{1}{c|}{} & \multicolumn{1}{c|}{} &  \\ \hline
\multicolumn{1}{|l|}{DoppelGANger} & \multicolumn{1}{c|}{$\times$} & \multicolumn{1}{c|}{$\times$} & \multicolumn{1}{c|}{$\times$} & \multicolumn{1}{c|}{$\times$} & \multicolumn{1}{c|}{} &  \\ \hline
\multicolumn{1}{|l|}{TVAE} & \multicolumn{1}{c|}{$\times$} & \multicolumn{1}{c|}{$\times$} & \multicolumn{1}{c|}{$\times$} & \multicolumn{1}{c|}{} & \multicolumn{1}{c|}{} &  \\ \hline
\multicolumn{1}{|l|}{TimeVAE} & \multicolumn{1}{c|}{$\times$} & \multicolumn{1}{c|}{$\times$} & \multicolumn{1}{c|}{$\times$} & \multicolumn{1}{c|}{} & \multicolumn{1}{c|}{} &  \\ \hline
\multicolumn{1}{|l|}{DP-VAEGM} & \multicolumn{1}{c|}{$\times$} & \multicolumn{1}{c|}{$\times$} & \multicolumn{1}{c|}{$\times$} & \multicolumn{1}{c|}{} & \multicolumn{1}{c|}{$\times$} &  \\ \hline
\multicolumn{1}{|l|}{TabDDPM} & \multicolumn{1}{c|}{$\times$} & \multicolumn{1}{c|}{$\times$} & \multicolumn{1}{c|}{$\times$} & \multicolumn{1}{c|}{$\times$} & \multicolumn{1}{c|}{} & $\times$ \\ \hline
\multicolumn{1}{|l|}{TSGM} & \multicolumn{1}{c|}{$\times$} & \multicolumn{1}{c|}{$\times$} & \multicolumn{1}{c|}{$\times$} & \multicolumn{1}{c|}{} & \multicolumn{1}{c|}{} &  \\ \hline
\multicolumn{1}{|l|}{SOS} & \multicolumn{1}{c|}{$\times$} & \multicolumn{1}{c|}{$\times$} & \multicolumn{1}{c|}{$\times$} & \multicolumn{1}{c|}{} & \multicolumn{1}{c|}{} &  \\ \hline
\multicolumn{1}{|l|}{STaSy} & \multicolumn{1}{c|}{$\times$} & \multicolumn{1}{c|}{$\times$} & \multicolumn{1}{c|}{$\times$} & \multicolumn{1}{c|}{} & \multicolumn{1}{c|}{} &  \\ \hline
\multicolumn{1}{|l|}{GOGGLE} & \multicolumn{1}{c|}{$\times$} & \multicolumn{1}{c|}{$\times$} & \multicolumn{1}{c|}{$\times$} & \multicolumn{1}{c|}{} & \multicolumn{1}{c|}{} &  \\ \hline
\multicolumn{1}{|l|}{GReaT} & \multicolumn{1}{c|}{$\times$} & \multicolumn{1}{c|}{$\times$} & \multicolumn{1}{c|}{$\times$} & \multicolumn{1}{c|}{} & \multicolumn{1}{c|}{} & $\times$ \\ \hline
\multicolumn{1}{|l|}{REaLTabFormer} & \multicolumn{1}{c|}{$\times$} & \multicolumn{1}{c|}{$\times$} & \multicolumn{1}{c|}{$\times$} & \multicolumn{1}{c|}{$\times$} & \multicolumn{1}{c|}{} & $\times$ \\ \hline
\multicolumn{1}{|l|}{TabuLa} & \multicolumn{1}{c|}{$\times$} & \multicolumn{1}{c|}{$\times$} & \multicolumn{1}{c|}{$\times$} & \multicolumn{1}{c|}{} & \multicolumn{1}{c|}{} & $\times$ \\ \hline
\multicolumn{1}{|l|}{KAMINO} & \multicolumn{1}{c|}{$\times$} & \multicolumn{1}{c|}{$\times$} & \multicolumn{1}{c|}{$\times$} & \multicolumn{1}{c|}{} & \multicolumn{1}{c|}{$\times$} &  \\ \hline
\multicolumn{1}{|l|}{AutoDiff} & \multicolumn{1}{c|}{$\times$} & \multicolumn{1}{c|}{$\times$} & \multicolumn{1}{c|}{$\times$} & \multicolumn{1}{c|}{$\times$} & \multicolumn{1}{c|}{} & $\times$ \\ \hline
\multicolumn{1}{|l|}{TabSyn} & \multicolumn{1}{c|}{$\times$} & \multicolumn{1}{c|}{$\times$} & \multicolumn{1}{c|}{$\times$} & \multicolumn{1}{c|}{$\times$} & \multicolumn{1}{c|}{} & $\times$ \\ \hline
\end{tabular}
}
\end{table*}

\begin{table*}[p]
\centering
\caption{Assessment of the \numberTools TDS tools on their reported performance on the functional requirements.}
\label{tab:matrix_req}
\scalebox{0.66}{
\begin{tabular}{l|cccccccccccc|}
\cline{2-13}
\multicolumn{1}{c|}{\textbf{}} &
  \multicolumn{12}{c|}{\textbf{Functional Requirements}} \\ \cline{2-13} 
\textbf{} &
  \multicolumn{6}{c|}{\textbf{Column Types}} &
  \multicolumn{1}{c|}{\multirow{2}{*}[-1em]{\textbf{\begin{tabular}[c]{@{}c@{}}Complex \\ Distributions\end{tabular}}}} &
  \multicolumn{2}{c|}{\textbf{Correlations}} &
  \multicolumn{2}{c|}{\textbf{Temporal}} &
  \multirow{2}{*}[-1em]{\textbf{\begin{tabular}[c]{@{}c@{}}Integrity \\ constraints\end{tabular}}} \\ \cline{1-7} \cline{9-12}
\multicolumn{1}{|l|}{\textbf{Tools}} &
  \multicolumn{1}{c|}{\textbf{Categorical}} &
  \multicolumn{1}{c|}{\textbf{\begin{tabular}[c]{@{}c@{}}Num. \\ Continuous\end{tabular}}} &
  \multicolumn{1}{c|}{\textbf{\begin{tabular}[c]{@{}c@{}}Num. \\ Discrete\end{tabular}}} &
  \multicolumn{1}{c|}{\textbf{Temporal}} &
  \multicolumn{1}{c|}{\textbf{Text}} &
  \multicolumn{1}{c|}{\textbf{\begin{tabular}[c]{@{}c@{}}Mixed\\ Cat./Num.\end{tabular}}} &
  \multicolumn{1}{c|}{} &
  \multicolumn{1}{c|}{\textbf{\begin{tabular}[c]{@{}c@{}}Intra-\\ table\end{tabular}}} &
  \multicolumn{1}{c|}{\textbf{\begin{tabular}[c]{@{}c@{}}Inter-\\ table\end{tabular}}} &
  \multicolumn{1}{c|}{\textbf{Short}} &
  \multicolumn{1}{c|}{\textbf{Long}} &
   \\ \hline
\multicolumn{1}{|l|}{All SMOTE} &
  \multicolumn{1}{c|}{} &
  \multicolumn{1}{c|}{$\times$} &
  \multicolumn{1}{c|}{} &
  \multicolumn{1}{c|}{} &
  \multicolumn{1}{c|}{} &
  \multicolumn{1}{c|}{} &
  \multicolumn{1}{c|}{} &
  \multicolumn{1}{c|}{} &
  \multicolumn{1}{c|}{} &
  \multicolumn{1}{c|}{} &
  \multicolumn{1}{c|}{} &
   \\ \hline
\multicolumn{1}{|l|}{ADASYN} &
  \multicolumn{1}{c|}{} &
  \multicolumn{1}{c|}{$\times$} &
  \multicolumn{1}{c|}{} &
  \multicolumn{1}{c|}{} &
  \multicolumn{1}{c|}{} &
  \multicolumn{1}{c|}{} &
  \multicolumn{1}{c|}{} &
  \multicolumn{1}{c|}{} &
  \multicolumn{1}{c|}{} &
  \multicolumn{1}{c|}{} &
  \multicolumn{1}{c|}{} &
   \\ \hline
\multicolumn{1}{|l|}{PrivBayes} &
  \multicolumn{1}{c|}{$\times$} &
  \multicolumn{1}{c|}{} &
  \multicolumn{1}{c|}{} &
  \multicolumn{1}{c|}{} &
  \multicolumn{1}{c|}{} &
  \multicolumn{1}{c|}{} &
  \multicolumn{1}{c|}{} &
  \multicolumn{1}{c|}{$\times$} &
  \multicolumn{1}{c|}{} &
  \multicolumn{1}{c|}{} &
  \multicolumn{1}{c|}{} &
   \\ \hline
\multicolumn{1}{|l|}{PrivMRF} &
  \multicolumn{1}{c|}{$\times$} &
  \multicolumn{1}{c|}{$\times$} &
  \multicolumn{1}{c|}{$\times$} &
  \multicolumn{1}{c|}{} &
  \multicolumn{1}{c|}{} &
  \multicolumn{1}{c|}{} &
  \multicolumn{1}{c|}{} &
  \multicolumn{1}{c|}{$\times$} &
  \multicolumn{1}{c|}{} &
  \multicolumn{1}{c|}{} &
  \multicolumn{1}{c|}{} &
   \\ \hline
\multicolumn{1}{|l|}{PrivLava} &
  \multicolumn{1}{c|}{$\times$} &
  \multicolumn{1}{c|}{$\times$} &
  \multicolumn{1}{c|}{$\times$} &
  \multicolumn{1}{c|}{} &
  \multicolumn{1}{c|}{} &
  \multicolumn{1}{c|}{} &
  \multicolumn{1}{c|}{$\times$} &
  \multicolumn{1}{c|}{$\times$} &
  \multicolumn{1}{c|}{$\times$} &
  \multicolumn{1}{c|}{} &
  \multicolumn{1}{c|}{} &
   \\ \hline
\multicolumn{1}{|l|}{medGAN} &
  \multicolumn{1}{c|}{$\times$} &
  \multicolumn{1}{c|}{$\times$} &
  \multicolumn{1}{c|}{$\times$} &
  \multicolumn{1}{c|}{} &
  \multicolumn{1}{c|}{} &
  \multicolumn{1}{c|}{} &
  \multicolumn{1}{c|}{} &
  \multicolumn{1}{c|}{$\times$} &
  \multicolumn{1}{c|}{} &
  \multicolumn{1}{c|}{} &
  \multicolumn{1}{c|}{} &
   \\ \hline
\multicolumn{1}{|l|}{PATE-GAN} &
  \multicolumn{1}{c|}{$\times$} &
  \multicolumn{1}{c|}{$\times$} &
  \multicolumn{1}{c|}{$\times$} &
  \multicolumn{1}{c|}{} &
  \multicolumn{1}{c|}{} &
  \multicolumn{1}{c|}{} &
  \multicolumn{1}{c|}{} &
  \multicolumn{1}{c|}{} &
  \multicolumn{1}{c|}{} &
  \multicolumn{1}{c|}{} &
  \multicolumn{1}{c|}{} &
   \\ \hline
\multicolumn{1}{|l|}{DP-GAN} &
  \multicolumn{1}{c|}{$\times$} &
  \multicolumn{1}{c|}{$\times$} &
  \multicolumn{1}{c|}{$\times$} &
  \multicolumn{1}{c|}{} &
  \multicolumn{1}{c|}{} &
  \multicolumn{1}{c|}{} &
  \multicolumn{1}{c|}{} &
  \multicolumn{1}{c|}{} &
  \multicolumn{1}{c|}{} &
  \multicolumn{1}{c|}{} &
  \multicolumn{1}{c|}{} &
   \\ \hline
\multicolumn{1}{|l|}{TableGAN} &
  \multicolumn{1}{c|}{$\times$} &
  \multicolumn{1}{c|}{$\times$} &
  \multicolumn{1}{c|}{$\times$} &
  \multicolumn{1}{c|}{} &
  \multicolumn{1}{c|}{} &
  \multicolumn{1}{c|}{} &
  \multicolumn{1}{c|}{} &
  \multicolumn{1}{c|}{$\times$} &
  \multicolumn{1}{c|}{} &
  \multicolumn{1}{c|}{} &
  \multicolumn{1}{c|}{} &
   \\ \hline
\multicolumn{1}{|l|}{TGAN} &
  \multicolumn{1}{c|}{$\times$} &
  \multicolumn{1}{c|}{$\times$} &
  \multicolumn{1}{c|}{$\times$} &
  \multicolumn{1}{c|}{} &
  \multicolumn{1}{c|}{} &
  \multicolumn{1}{c|}{} &
  \multicolumn{1}{c|}{} &
  \multicolumn{1}{c|}{$\times$} &
  \multicolumn{1}{c|}{} &
  \multicolumn{1}{c|}{} &
  \multicolumn{1}{c|}{} &
   \\ \hline
\multicolumn{1}{|l|}{DTGAN} &
  \multicolumn{1}{c|}{$\times$} &
  \multicolumn{1}{c|}{$\times$} &
  \multicolumn{1}{c|}{$\times$} &
  \multicolumn{1}{c|}{} &
  \multicolumn{1}{c|}{} &
  \multicolumn{1}{c|}{} &
  \multicolumn{1}{c|}{} &
  \multicolumn{1}{c|}{$\times$} &
  \multicolumn{1}{c|}{} &
  \multicolumn{1}{c|}{} &
  \multicolumn{1}{c|}{} &
   \\ \hline
\multicolumn{1}{|l|}{CTGAN} &
  \multicolumn{1}{c|}{$\times$} &
  \multicolumn{1}{c|}{$\times$} &
  \multicolumn{1}{c|}{$\times$} &
  \multicolumn{1}{c|}{} &
  \multicolumn{1}{c|}{} &
  \multicolumn{1}{c|}{} &
  \multicolumn{1}{c|}{$\times$} &
  \multicolumn{1}{c|}{$\times$} &
  \multicolumn{1}{c|}{} &
  \multicolumn{1}{c|}{} &
  \multicolumn{1}{c|}{} &
   \\ \hline
   \multicolumn{1}{|l|}{C3TGAN} &
  \multicolumn{1}{c|}{$\times$} &
  \multicolumn{1}{c|}{$\times$} &
  \multicolumn{1}{c|}{$\times$} &
  \multicolumn{1}{c|}{} &
  \multicolumn{1}{c|}{} &
  \multicolumn{1}{c|}{} &
  \multicolumn{1}{c|}{} &
  \multicolumn{1}{c|}{$\times$} &
  \multicolumn{1}{c|}{} &
  \multicolumn{1}{c|}{} &
  \multicolumn{1}{c|}{} &
  $\times$ \\ \hline
\multicolumn{1}{|l|}{CTAB-GAN} &
  \multicolumn{1}{c|}{$\times$} &
  \multicolumn{1}{c|}{$\times$} &
  \multicolumn{1}{c|}{$\times$} &
  \multicolumn{1}{c|}{} &
  \multicolumn{1}{c|}{} &
  \multicolumn{1}{c|}{$\times$} &
  \multicolumn{1}{c|}{$\times$} &
  \multicolumn{1}{c|}{$\times$} &
  \multicolumn{1}{c|}{} &
  \multicolumn{1}{c|}{} &
  \multicolumn{1}{c|}{} &
   \\ \hline
\multicolumn{1}{|l|}{CTAB-GAN+} &
  \multicolumn{1}{c|}{$\times$} &
  \multicolumn{1}{c|}{$\times$} &
  \multicolumn{1}{c|}{$\times$} &
  \multicolumn{1}{c|}{} &
  \multicolumn{1}{c|}{} &
  \multicolumn{1}{c|}{$\times$} &
  \multicolumn{1}{c|}{$\times$} &
  \multicolumn{1}{c|}{$\times$} &
  \multicolumn{1}{c|}{} &
  \multicolumn{1}{c|}{} &
  \multicolumn{1}{c|}{} &
   \\ \hline
\multicolumn{1}{|l|}{GANBLR} &
  \multicolumn{1}{c|}{$\times$} &
  \multicolumn{1}{c|}{} &
  \multicolumn{1}{c|}{} &
  \multicolumn{1}{c|}{} &
  \multicolumn{1}{c|}{} &
  \multicolumn{1}{c|}{} &
  \multicolumn{1}{c|}{$\times$} &
  \multicolumn{1}{c|}{$\times$} &
  \multicolumn{1}{c|}{} &
  \multicolumn{1}{c|}{} &
  \multicolumn{1}{c|}{} &
   \\ \hline
\multicolumn{1}{|l|}{GANBLR+} &
  \multicolumn{1}{c|}{$\times$} &
  \multicolumn{1}{c|}{$\times$} &
  \multicolumn{1}{c|}{$\times$} &
  \multicolumn{1}{c|}{} &
  \multicolumn{1}{c|}{} &
  \multicolumn{1}{c|}{} &
  \multicolumn{1}{c|}{$\times$} &
  \multicolumn{1}{c|}{$\times$} &
  \multicolumn{1}{c|}{} &
  \multicolumn{1}{c|}{} &
  \multicolumn{1}{c|}{} &
   \\ \hline
\multicolumn{1}{|l|}{TimeGAN} &
  \multicolumn{1}{c|}{$\times$} &
  \multicolumn{1}{c|}{$\times$} &
  \multicolumn{1}{c|}{$\times$} &
  \multicolumn{1}{c|}{$\times$} &
  \multicolumn{1}{c|}{} &
  \multicolumn{1}{c|}{} &
  \multicolumn{1}{c|}{} &
  \multicolumn{1}{c|}{$\times$} &
  \multicolumn{1}{c|}{} &
  \multicolumn{1}{c|}{$\times$} &
  \multicolumn{1}{c|}{} &
   \\ \hline
\multicolumn{1}{|l|}{DoppelGANger} &
  \multicolumn{1}{c|}{$\times$} &
  \multicolumn{1}{c|}{$\times$} &
  \multicolumn{1}{c|}{$\times$} &
  \multicolumn{1}{c|}{$\times$} &
  \multicolumn{1}{c|}{} &
  \multicolumn{1}{c|}{} &
  \multicolumn{1}{c|}{$\times$} &
  \multicolumn{1}{c|}{$\times$} &
  \multicolumn{1}{c|}{} &
  \multicolumn{1}{c|}{$\times$} &
  \multicolumn{1}{c|}{$\times$} &
   \\ \hline
\multicolumn{1}{|l|}{TVAE} &
  \multicolumn{1}{c|}{$\times$} &
  \multicolumn{1}{c|}{$\times$} &
  \multicolumn{1}{c|}{$\times$} &
  \multicolumn{1}{c|}{} &
  \multicolumn{1}{c|}{} &
  \multicolumn{1}{c|}{} &
  \multicolumn{1}{c|}{$\times$} &
  \multicolumn{1}{c|}{$\times$} &
  \multicolumn{1}{c|}{} &
  \multicolumn{1}{c|}{} &
  \multicolumn{1}{c|}{} &
   \\ \hline
\multicolumn{1}{|l|}{TimeVAE} &
  \multicolumn{1}{c|}{$\times$} &
  \multicolumn{1}{c|}{$\times$} &
  \multicolumn{1}{c|}{$\times$} &
  \multicolumn{1}{c|}{$\times$} &
  \multicolumn{1}{c|}{} &
  \multicolumn{1}{c|}{} &
  \multicolumn{1}{c|}{} &
  \multicolumn{1}{c|}{$\times$} &
  \multicolumn{1}{c|}{} &
  \multicolumn{1}{c|}{$\times$} &
  \multicolumn{1}{c|}{} &
   \\ \hline
\multicolumn{1}{|l|}{DP-VAE-GM} &
  \multicolumn{1}{c|}{$\times$} &
  \multicolumn{1}{c|}{$\times$} &
  \multicolumn{1}{c|}{$\times$} &
  \multicolumn{1}{c|}{} &
  \multicolumn{1}{c|}{} &
  \multicolumn{1}{c|}{} &
  \multicolumn{1}{c|}{} &
  \multicolumn{1}{c|}{$\times$} &
  \multicolumn{1}{c|}{} &
  \multicolumn{1}{c|}{} &
  \multicolumn{1}{c|}{} &
   \\ \hline
\multicolumn{1}{|l|}{TabDDPM} &
  \multicolumn{1}{c|}{$\times$} &
  \multicolumn{1}{c|}{$\times$} &
  \multicolumn{1}{c|}{$\times$} &
  \multicolumn{1}{c|}{} &
  \multicolumn{1}{c|}{} &
  \multicolumn{1}{c|}{} &
  \multicolumn{1}{c|}{$\times$} &
  \multicolumn{1}{c|}{$\times$} &
  \multicolumn{1}{c|}{} &
  \multicolumn{1}{c|}{} &
  \multicolumn{1}{c|}{} &
   \\ \hline
\multicolumn{1}{|l|}{TSGM} &
  \multicolumn{1}{c|}{$\times$} &
  \multicolumn{1}{c|}{$\times$} &
  \multicolumn{1}{c|}{$\times$} &
  \multicolumn{1}{c|}{$\times$} &
  \multicolumn{1}{c|}{} &
  \multicolumn{1}{c|}{} &
  \multicolumn{1}{c|}{} &
  \multicolumn{1}{c|}{$\times$} &
  \multicolumn{1}{c|}{} &
  \multicolumn{1}{c|}{$\times$} &
  \multicolumn{1}{c|}{} &
   \\ \hline
\multicolumn{1}{|l|}{SOS} &
  \multicolumn{1}{c|}{$\times$} &
  \multicolumn{1}{c|}{$\times$} &
  \multicolumn{1}{c|}{$\times$} &
  \multicolumn{1}{c|}{} &
  \multicolumn{1}{c|}{} &
  \multicolumn{1}{c|}{} &
  \multicolumn{1}{c|}{$\times$} &
  \multicolumn{1}{c|}{$\times$} &
  \multicolumn{1}{c|}{} &
  \multicolumn{1}{c|}{} &
  \multicolumn{1}{c|}{} &
   \\ \hline
\multicolumn{1}{|l|}{STaSy} &
  \multicolumn{1}{c|}{$\times$} &
  \multicolumn{1}{c|}{$\times$} &
  \multicolumn{1}{c|}{$\times$} &
  \multicolumn{1}{c|}{} &
  \multicolumn{1}{c|}{} &
  \multicolumn{1}{c|}{} &
  \multicolumn{1}{c|}{$\times$} &
  \multicolumn{1}{c|}{$\times$} &
  \multicolumn{1}{c|}{} &
  \multicolumn{1}{c|}{} &
  \multicolumn{1}{c|}{} &
   \\ \hline
\multicolumn{1}{|l|}{GOGGLE} &
  \multicolumn{1}{c|}{$\times$} &
  \multicolumn{1}{c|}{$\times$} &
  \multicolumn{1}{c|}{$\times$} &
  \multicolumn{1}{c|}{} &
  \multicolumn{1}{c|}{} &
  \multicolumn{1}{c|}{} &
  \multicolumn{1}{c|}{$\times$} &
  \multicolumn{1}{c|}{$\times$} &
  \multicolumn{1}{c|}{} &
  \multicolumn{1}{c|}{} &
  \multicolumn{1}{c|}{} &
   \\ \hline
\multicolumn{1}{|l|}{GReaT} &
  \multicolumn{1}{c|}{$\times$} &
  \multicolumn{1}{c|}{$\times$} &
  \multicolumn{1}{c|}{$\times$} &
  \multicolumn{1}{c|}{} &
  \multicolumn{1}{c|}{$\times$} &
  \multicolumn{1}{c|}{$\times$} &
  \multicolumn{1}{c|}{$\times$} &
  \multicolumn{1}{c|}{$\times$} &
  \multicolumn{1}{c|}{} &
  \multicolumn{1}{c|}{} &
  \multicolumn{1}{c|}{} &
   \\ \hline
\multicolumn{1}{|l|}{REaLTabFormer} &
  \multicolumn{1}{c|}{$\times$} &
  \multicolumn{1}{c|}{$\times$} &
  \multicolumn{1}{c|}{$\times$} &
  \multicolumn{1}{c|}{$\times$} &
  \multicolumn{1}{c|}{$\times$} &
  \multicolumn{1}{c|}{$\times$} &
  \multicolumn{1}{c|}{$\times$} &
  \multicolumn{1}{c|}{$\times$} &
  \multicolumn{1}{c|}{$\times$} &
  \multicolumn{1}{c|}{} &
  \multicolumn{1}{c|}{} &
   \\ \hline
\multicolumn{1}{|l|}{TabuLa} &
  \multicolumn{1}{c|}{$\times$} &
  \multicolumn{1}{c|}{$\times$} &
  \multicolumn{1}{c|}{$\times$} &
  \multicolumn{1}{c|}{} &
  \multicolumn{1}{c|}{$\times$} &
  \multicolumn{1}{c|}{$\times$} &
  \multicolumn{1}{c|}{$\times$} &
  \multicolumn{1}{c|}{$\times$} &
  \multicolumn{1}{c|}{} &
  \multicolumn{1}{c|}{} &
  \multicolumn{1}{c|}{} &
   \\ \hline
\multicolumn{1}{|l|}{KAMINO} &
  \multicolumn{1}{c|}{$\times$} &
  \multicolumn{1}{c|}{$\times$} &
  \multicolumn{1}{c|}{$\times$} &
  \multicolumn{1}{c|}{} &
  \multicolumn{1}{c|}{} &
  \multicolumn{1}{c|}{$\times$} &
  \multicolumn{1}{c|}{} &
  \multicolumn{1}{c|}{$\times$} &
  \multicolumn{1}{c|}{} &
  \multicolumn{1}{c|}{} &
  \multicolumn{1}{c|}{} &
  $\times$ \\ \hline
\multicolumn{1}{|l|}{AutoDiff} &
  \multicolumn{1}{c|}{$\times$} &
  \multicolumn{1}{c|}{$\times$} &
  \multicolumn{1}{c|}{$\times$} &
  \multicolumn{1}{c|}{} &
  \multicolumn{1}{c|}{} &
  \multicolumn{1}{c|}{$\times$} &
  \multicolumn{1}{c|}{$\times$} &
  \multicolumn{1}{c|}{$\times$} &
  \multicolumn{1}{c|}{} &
  \multicolumn{1}{c|}{} &
  \multicolumn{1}{c|}{} &
   \\ \hline
\multicolumn{1}{|l|}{TabSyn} &
  \multicolumn{1}{c|}{$\times$} &
  \multicolumn{1}{c|}{$\times$} &
  \multicolumn{1}{c|}{$\times$} &
  \multicolumn{1}{c|}{} &
  \multicolumn{1}{c|}{} &
  \multicolumn{1}{c|}{$\times$} &
  \multicolumn{1}{c|}{$\times$} &
  \multicolumn{1}{c|}{$\times$} &
  \multicolumn{1}{c|}{} &
  \multicolumn{1}{c|}{} &
  \multicolumn{1}{c|}{} &
   \\ \hline
\end{tabular}
}
\end{table*}

\begin{figure*}[p]
    \centering
    \includegraphics[height=0.6\textheight, angle=270]{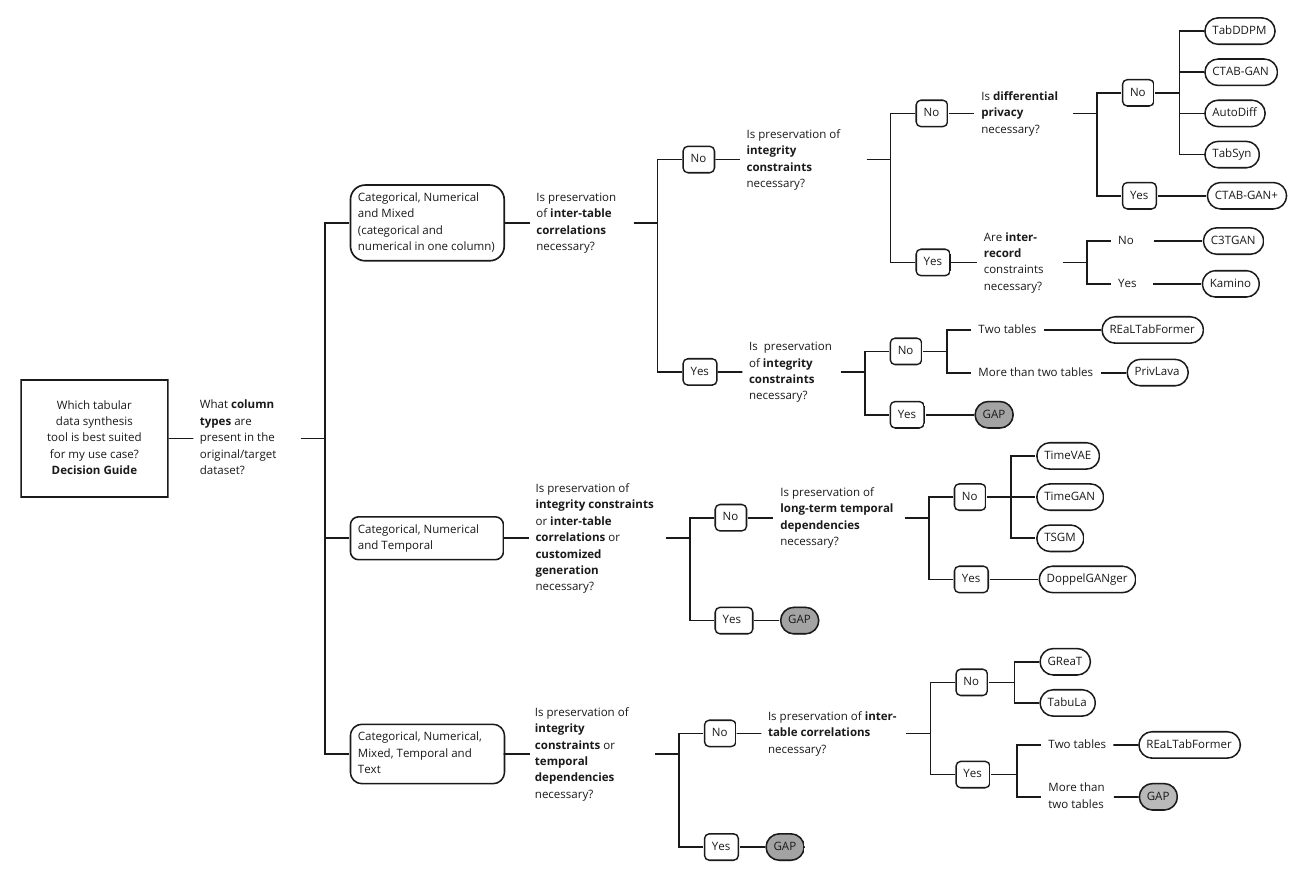}
    \Description[<Decision Guide>]{<Decision guide resulting from the assessment.>}
    \caption{Decision guide resulting from the assessment of \numberTools TDS tools, based on their reported performance on the functional requirements identified in our work.}
    \label{img:guide}
\end{figure*}

The first step in our assessment was determining which purposes were addressed by each TDS tool. For example, we consider TDS tools suitable for privacy protection if they include privacy-preserving techniques in their algorithms, as well as privacy evaluation metrics in their experiments. All the tools in our decision guide are at least able to address missing value imputation, dataset balancing, and data augmentation. The difference between the tools lies in whether they are suitable for privacy (either simple\footnote{We consider a privacy-preserving technique to be simple if it cannot provide a concrete privacy guarantee, as in the case of differential privacy, but it ensures that the generated records do not exactly resemble the original records, which is usually evaluated by measuring their distance (see DCR in Table~\ref{tab: Evaluation Metrics}).} or differential privacy \cite{Dwork2006}), whether they allow customized generation, and whether they meet the different requirements.

Therefore, in the second step we assessed the TDS tools' reported performance on functional requirements. All TDS approaches included here are able to synthesize multiple dependent columns. Furthermore, we marked the column types which are reported to be effectively handled by the tool, based on the datasets used in their experiments. Some approaches only included datasets with either categorical or numerical columns, however, the most complete approaches are also able to work with temporal, text, and mixed column types.

Furthermore, we marked whether the TDS tools are reported to perform well with complex distributions or whether they assumed columns to follow a Gaussian distribution. Some oldest tools mainly address privacy and do not aim to preserve correlations between columns. Therefore, we also marked whether or not the tools preserve column correlations. Similarly, for tools specially designed to handle time series datasets, we marked whether they address short-term and/or long-term dependencies.

Finally, we assessed whether the TDS tools address challenges outside the ML community, such as integrity constraints and inter-table correlations. For inter-table correlations, we checked whether the tools address two (e.g., a parent-child relationship) or more tables. 

An assessment of the non-functional requirements was not possible because there is not enough information published to allow a fair and comprehensive comparison of the required customization, pre-processing, and hardware requirements as well as the required run-time and memory consumption. 

The complete assessment was then used to consolidate all of these factors into a decision guide, which is shown in Figure \ref{img:guide} in the shape of a decision tree. The complete decision guide with all \numberTools tools and all decision factors is too big for printed representation. For this reason, we excluded the simplest branches from Figure \ref{img:guide}, which include: 
\begin{inparaenum}[(i)]
    \item TDS tools that can only be used to address the purposes of missing value imputation, dataset balancing, or data augmentation,
    \item TDS tools that can handle only categorical or numerical columns,
    \item TDS tools that do not effectively preserve column correlations, and
    \item TDS tools that assume all column distributions to be Gaussian.
\end{inparaenum}

The decision guide includes five node levels, ending with leaves which are either a suitable TDS tool for the use case or a "Gap-Leaf", representing a research gap. The questions included in the guide are meant to show users the differences between the TDS tools. With the limited amount of information available regarding the tools' resource efficiency, the guide suggests tools that cover the requirements of the corresponding branch and avoids recommending overly complex tools. 

This decision guide allows users to assess the suitability of a TDS tool for their use case by answering questions about their own dataset and the intended purpose of the data synthesis. Since it requires little to no expertise about the tools' underlying models (e.g., GANs, transformers, or probabilistic diffusion), we think this approach is much simpler than navigating the decision process using the differences of the individual TDS models. If the decision process ends up with a "Gap-Leaf", the users can still identify the nearest possible tools and their limitations. For example, if their dataset includes categorical, numerical, temporal, and text columns, and the preservation of integrity constraints is necessary, the users will find that there are no suitable TDS tools to the day. However, they can see that REaLTabFormer~\cite{REaLTabFormer} works for all those column types, but does not necessarily preserve integrity constraints and that KAMINO~\cite{Kamino} preserves integrity constraints,
but does not support temporal and text columns.

The most significant research gaps that we have identified in our assessment are: 
\begin{itemize}
    \item TDS tools that effectively preserve integrity constraints, while handling all column types and complex column distributions,
    \item TDS tools that preserve inter-table correlations between columns, especially for datasets containing more than two tables, while handling complex column types, such as temporal or text, 
    \item TDS tools that preserve temporal dependencies between columns, while addressing all column types and complex column distributions, and
    \item A universal TDS tool that addresses all functional requirements.
\end{itemize}

During our assessment, we observed a lack of information on non-functional requirements in many publications, such as the level of customization, pre-processing, and hardware required, as well as the resource efficiency and scalability of the tool.

\section{Conclusion}
\label{sec:conclusion}
Data scarcity and data privacy have become fundamental problems for data-driven models across application domains. 
While data synthesis tools are already used to mitigate this issue, choosing the right tool for a given context has become increasingly complex. 

In this paper, we provide an overview of the challenges and solutions that currently exist in the field of tabular data synthesis (TDS). 
To this end, we compiled both the relevant requirements on the side of the application as well as their interaction with the capabilities and limitations exhibited by concrete TDS tools. 
We evaluated \numberTools TDS tools with respect to these requirements resulting in two assessment matrices.
Based on the matrices, we developed a decision guide that supports users in selecting the right tool for their specific use case. 

From the four research gaps identified during this study, we place special emphasis on the fact that TDS research has so far focused on the ML community. We believe further research is needed to expand TDS tools or combinations of methods that ensure the preservation of integrity constraints, or the capability of generating datasets with complex schemas consisting of multiple tables.

Building on the work presented here, future work includes automatically assessing the tools suitability for different real-world applications. Our long-term goal is designing a benchmarking framework that allows evaluating TDS tools on their fitness for use of diverse applications, on the basis of which users will be guided by experimental results rather than reported performance numbers like in the current iteration of our decision guide.

\begin{acks}
This research was partially funded by the Deutsche Forschungsgemeinschaft (DFG, German Research Foundation) – Project Number 495170629 and Federal Ministry for Economic Affairs and Climate Action (BMWK) - Project Number 01MD22001D.
\end{acks}

\balance
\bibliographystyle{ACM-Reference-Format}
\bibliography{mybib}

\end{document}